\documentclass{article}

\usepackage{arxiv}

\usepackage[utf8]{inputenc} 
\usepackage[T1]{fontenc}    
\usepackage{hyperref}       
\usepackage{url}            
\usepackage{booktabs}       
\usepackage{amsfonts}       
\usepackage{nicefrac}       
\usepackage{microtype}      
\usepackage{graphicx}
\usepackage{subcaption}
\usepackage{mathrsfs}
\usepackage{amsmath}
\usepackage[ruled,vlined]{algorithm2e}

\title{Probabilistic prediction of the heave motions of a semi-submersible by a deep learning problem model}

\date{Oct 8, 2021}	

\author{ {Xiaoxian Guo} {, Xiantao Zhang, } {Xinliang Tian, } {Wenyue Lu, } {Xin Li}\\
		State Key Laboratory of Ocean Engineering\\
		Shanghai Jiao Tong University\\
		Shanghai 200240, China \\
	}

\renewcommand{\headeright}{}
\renewcommand{\undertitle}{}

\begin{document}
\maketitle

\begin{abstract}
	The real-time motion prediction of a floating offshore platform refers to forecasting its motions in the following one- or two-wave cycles, which helps improve the performance of a motion compensation system and provides useful early warning information. In this study, we extend a deep learning (DL) model, which could predict the heave and surge motions of a floating semi-submersible 20 to 50 seconds ahead with good accuracy, to quantify its uncertainty of the predictive time series with the help of the dropout technique. By repeating the inference several times, it is found that the collection of the predictive time series is a Gaussian process (GP). The DL model with dropout learned a kernel inside, and the learning procedure was similar to GP regression. Adding noise into training data could help the model to learn more robust features from the training data, thereby leading to a better performance on test data with a wide noise level range. This study extends the understanding of the DL model to predict the wave excited motions of an offshore platform. 
\end{abstract}

\keywords{Real-time motion prediction \and confidence interval \and dropout \and LSTM \and semi-submersible}

\section{Introduction}
Ships or offshore floating platforms move in six degrees of freedom (DOFs) excited by environmental forces, such as wave, wind, or ocean current. The six DOF motions constitute a remarkable challenge to carry out works offshore, thereby limiting the abilities of many motion-sensitive operations. The real-time motion prediction of a vessel or a floating platform usually refers to forecasting its motions in the following tens of seconds (one- or two-wave cycles) in real time. It can help to improve the performance of a motion compensation system and also provide useful early warning information. An aircraft landing on a carrier or a missile or rocket launching from a vessel also needs real-time motion prediction for its better performance. 

The existing models for real-time motion prediction fall into two categories: a) a model combining the ocean wave kinetics and ship dynamics to predict the ship behavior in the future; b) a model purely relying on direct time series analysis. The Kalman filter technique has been used for vessel motion prediction since the 1980s~\cite{Triantafyllou1981,Triantafyllou1983}. It requires a state-space model of the vessel based on a full knowledge of hydrodynamics. Given that the added mass, damping, and wave excitation forces are frequency-dependent, the estimation of the peak frequency of the wave spectrum becomes crucial. Yumori~\cite{Yumori1981} demonstrated an auto-regressive moving average model to predict the ship behavior approximately 2 to 4 seconds ahead. Nielsen et al.~\cite{Nielsen2018} proposed a model based on auto-correction function to predict the response sequence 15 to 60 seconds ahead of the current time. Direct time series analysis, including the auto-regressive or auto-correction function model, does not require prior knowledge about the ship responses. Other models based on direct time series analysis can be found in \cite{Broome1990,Broome1998,Broome1998a,Zhao2004,Duan2015}.

Deep learning (DL) is a powerful tool to extract complex mapping between input and output purely based on the existing labeled data. The trainable parameters are learned by minimizing the loss that is defined between the prediction and ground truth through a back propagation algorithm. Khan et al.~\cite{Khan2005} achieved roll motion prediction extending up to 7 seconds using a three-layer fully connected (FC) neural network (NN). Li et al.~\cite{Li2019} used an NN with two hidden layers to predict a wave excitation force extending 2.5 seconds ahead for the controller of a wave energy converter. Recurrent NNs (RNNs) were designed to consider the inherent order of the data and were not subjected to a fixed input data length. Notably, RNNs are highly successful in natural language processing. Presently, the so-called long short-term memory (LSTM) \cite{Hochreiter1997} algorithms are the most popular type of RNNs, which also show their feasibility for ship motion prediction with good accuracy, as reported in~\cite{Hua2019,Duan2019, Ferrandis2019}. 

All the above-mentioned methods or models give deterministic predictions without the knowledge of prediction confidence or uncertainty. Uncertainty can be categorized into aleatoric and epistemic uncertainty. Aleatoric uncertainty comes from noises in input data, and it is usually reducible. Epistemic uncertainty refers to the uncertainty in model parameters and structure, and it is inherent and irreducible~\cite{Gal2016a}. For ship motion prediction, the uncertainty of the prediction comes from both noisy input measurements and parameters and the structure of the model. In practice, quantifying uncertainty is equally important to the deterministic prediction itself for the decision-making process. Knowing the uncertainty information of the prediction, for example, underconfidence or false overconfidence, can also help to improve the performance of the prediction model. 

The uncertainty or confidence of the prediction made by the DL model was originally discussed for artificial intelligence safety of medical diagnosis models, such as auto diagnostics for magnetic resonance imaging scans~\cite{Bernard2018, Lundervold2019}. Gal~\cite{Gal2016a} studied DL uncertainty, and proposed that dropout could be used to approximate the uncertainty of the DL model~\cite{Gal2016}. The dropout layer widely exists in many DL models for preventing over-fitting, and therefore, it is very practical for estimating the prediction uncertainty of existing models. A successful application can be found in \cite{Zhu2017} for the probabilistic prediction of the number of trips for Uber use. Another approach is the Bayesian DL model that treats the learnable parameters (weights and bias) as random variables instead of particular values~\cite{Blundell2015}, which is much more complicated compared to the approach involving dropout. Bayesian LSTM models for time series forecasting can be found in \cite{Zhang2020,Cabras2020}. Gaussian process regression (GPR) is also a good choice for function approximation with uncertainty estimation as it is flexible, robust to overfitting, and provides well-calibrated predictive uncertainty. However, in practice, GPR has poor accuracy for exploitation problems.

The DL model for real-time motion prediction proposed by Guo et al.~\cite{Guo2021} will be the starting point of this paper. By using this DL model, we predicted the heave and surge motions of a semi-submersible 20 to 50 seconds into future with good accuracy. In this study, we quantified the uncertainty of the prediction by inserting dropout layers into this model. Thereafter, we discussed the performance of the model and the properties of the predictive time series. Finally, we added noises into both the training and test data to investigate the model performance. 

The remainder of this paper is organized as follows: in the next section, the data used for learning were discussed at first, including where the data came from and how to process them into input--output pairs for the learning procedure. In \S~\ref{sec:learningmodel}, we discussed the structure of the DL model and how to obtain the uncertainty of the predictions with the dropout technique. Finally, in \S~\ref{sec:results}, the results were presented and discussed.

\section{Datasets}\label{sec:datasets}
\subsection{Model tests in offshore basin}
As the real offshore platform for the oil and gas industry is so huge (approximately hundreds of thousands of tons) that we cannot perform direct tests with the real one, a scaled model test becomes a common technique to evaluate the dynamic performance of an offshore platform. We usually perform the model test in a wave basin. A scaled model is moored in the basin, and the model’s motion responses are observed under generated environments, including wave, current, and wind. Thereafter, the results are scaled up to the prototype based on Froude’s scaling law, thereby reflecting the real situation.

\begin{figure}[th]
	\centering
	\includegraphics[width=0.95\linewidth]{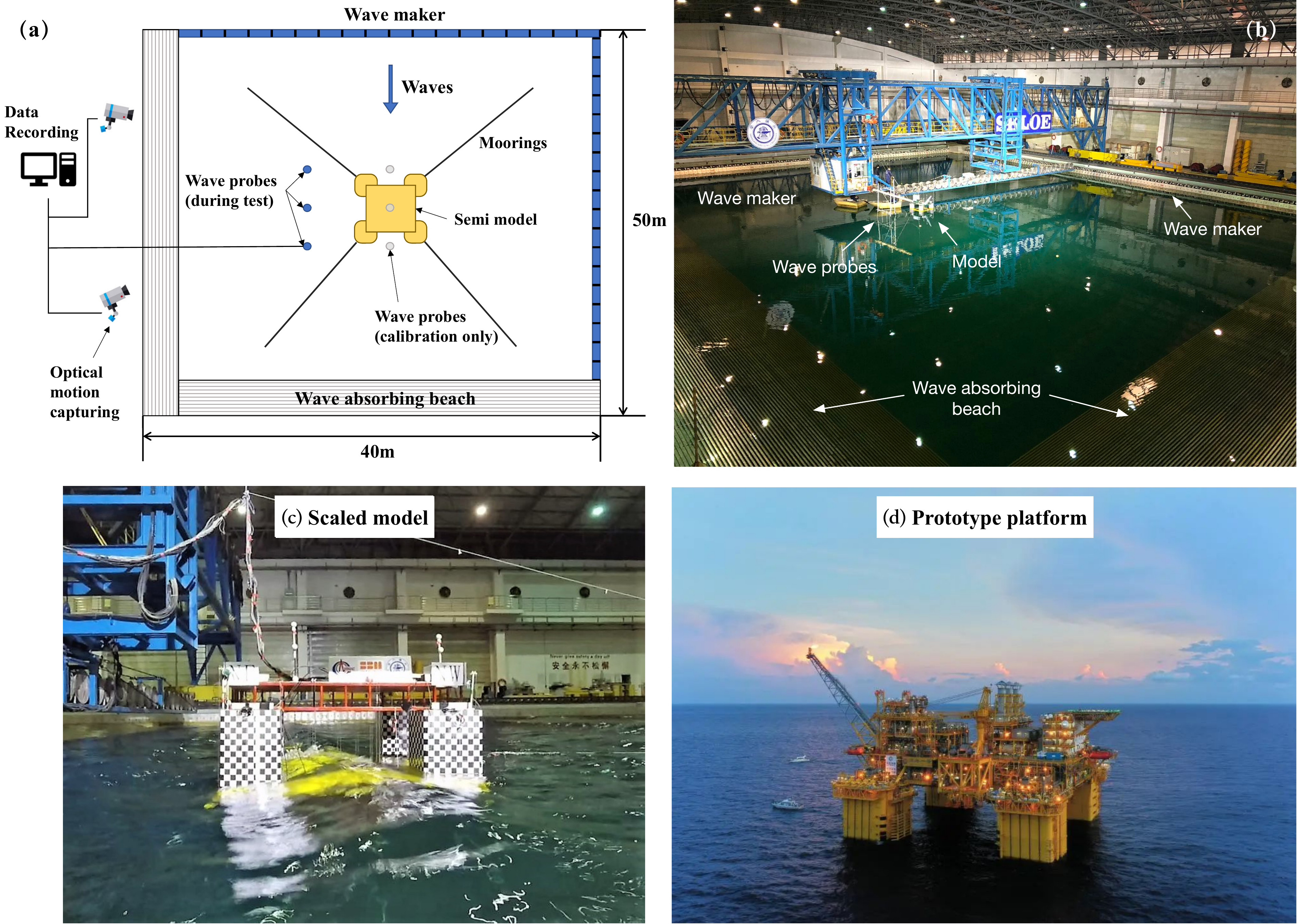}
	\caption{(a) Top view of the semi-model arrangement in the deep water wave basin; (b) picture of the wave basin; (c) picture of the model during testing in waves; and (d) the prototype platform, which is active at the Lingshui 17-2 gas field, South China Sea (from Internet: \url{https://www.xindemarinenews.com/china/30567.html}).}
	\label{fig:basinview}
\end{figure}

We performed this test for a semi-submersible in the deep water wave basin at Shanghai Jiao Tong University (SJTU), China. This wave basin is 50.0 m in length, 40.0 m in width, and up to 10.0 m in depth, with a large-area movable bottom. The L-shaped wave maker, which spans two adjacent sides, was equipped to generate the ocean waves in the basin. The passive wave-absorbing beach was also equipped with an optimized parabola profile and damping grids for wave dissipation. 

The testing semi-submersible, as shown in Fig.~\ref{fig:basinview} (c), is a typical platform for offshore oil and gas exploitation. The real one is designed for the Lingshui 17-2 gas field, South China Sea, as seen in Fig.~\ref{fig:basinview} (d). The scale ratio was set to 1:60. The model was made out of wood and adjusted on the trimming table to ensure its correct properties, including mass, center of gravity, and radius of gyration. The model was moored in the wave basin by catenary mooring lines attached to the model at corners, which provided a restoring force to maintain the position of the model. An optical motion capture system provided by Qualisys was used for capturing the six DOF motions at the center of the waterline area of the model, and three wave probes were installed beside the model to record the wave elevations.

\begin{figure}[ht]
	\centering
	\includegraphics[width=\linewidth]{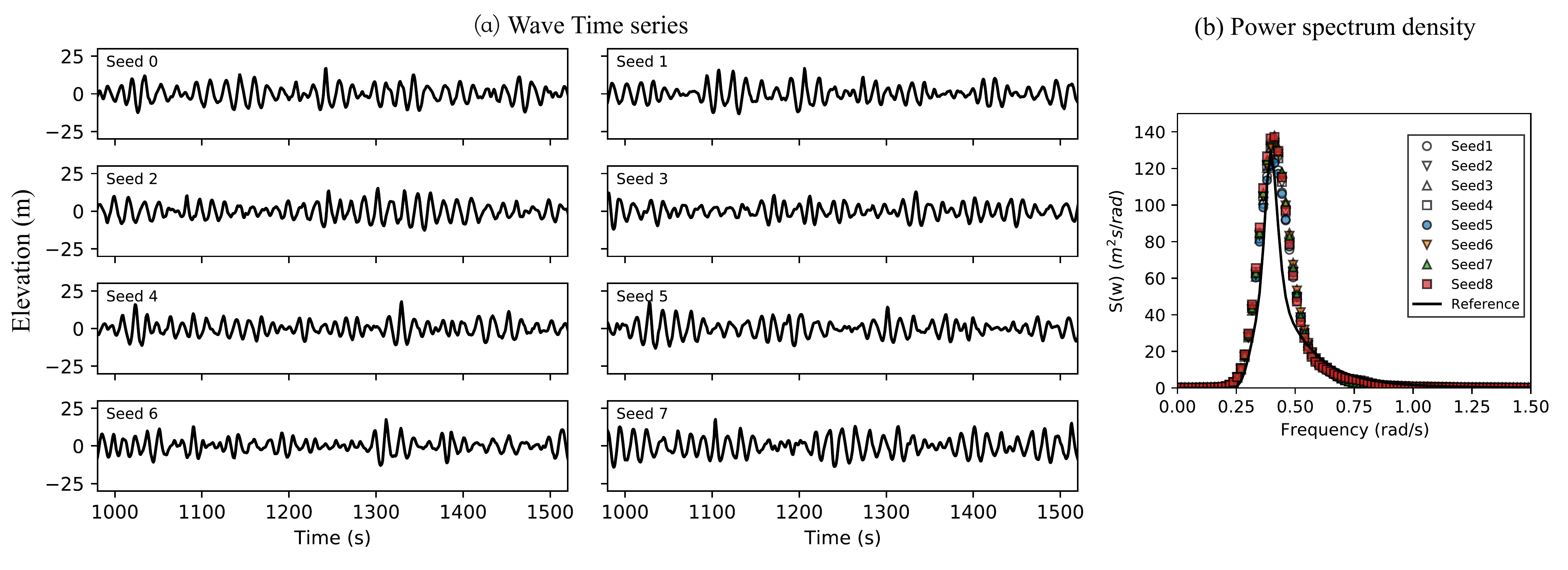}
	\caption{(a) The eight wave time series with different random seeds and (b) the corresponding power spectrum densities ($H_s=$17.4m and $T_p=$15.9s).}
	\label{fig:wavecases}
\end{figure}

Only waves were considered as environmental excitations in the present study. The irregular waves generated follow the JONSWAP spectrum, which is defined thus:
\begin{equation}\label{equ:jonswap}
	S(\omega)= \alpha H_s^2 \frac{\omega^{-5}}{\omega_p^{-4}} \exp\big[-1.25(\omega/\omega_p)^{-4}\big]
	\gamma^{\exp\big[-\frac{(\omega-\omega_p)^2}{2\tau^2\omega_p^2}\big]},
\end{equation}
where $\gamma$ is the peakedness parameter ($\gamma=2.4$ throughout this study), $\tau$ is the shape parameter ($\tau=0.07$ for $\omega>\omega_p$ and $\tau=0.09$ for $\omega<\omega_p$, as suggested in~\cite{DNVRP2010}), $H_s$ is the significant wave height, and $\omega_p=2\pi/T_p$ is the angular spectral peak frequency. In this study, two different combinations of $H_s$ and $T_p$ were considered.

We generated the wave time series as a sum of a collection of sinusoidal wave components,
\begin{equation}
	\eta(t) = \sum_i^N \eta_i(t) = \sum_i^N \zeta_i \cos{(\omega_i t + \epsilon_i)},
\end{equation}
where $\zeta_i$ is the amplitude decided by the wave spectrum ($\zeta_i = \sqrt{2 S(\omega_i) \Delta \omega}$), and $\epsilon_i$ is a random phase angle. The collection of random phase angles $[\epsilon_1, \epsilon_2, ..., \epsilon_i, ...]$ is the random seed. The predefined amplitude of each wave component ensures that the generated wave time series follows the expected JONSWAP spectrum in the frequency domain, and the random phase provides randomness in the time series. With different random seeds, we got different wave time series (see fig.~\ref{fig:wavecases} (a)) but all the time series had the same spectrum shape in the frequency domain (see fig.~\ref{fig:wavecases} (b)). Here, for each combination of $H_s$ and $T_p$, eight different seeds were used. The specified wave conditions were calibrated before the model test. Instead of the model, three wave probes were placed at the center of the basin to record the wave elevation at those points for wave calibration. 

For each case, we started wave generation at first, and then synchronously recorded the wave elevation and motions of the model at 10 Hz for a 30-minute period in model scale. Note that all the results were scaled up to a prototype with a scale ratio of 60. The sampling rate is 1.29 Hz at prototype. For each test case, the first 2-minute time series was removed from the results to eliminate the setting-up effects. As seen in Fig.~\ref{fig:wavecases}, the time series was stationary GP. Notably, that the heave motion of the offshore platform excited by the present wave was also stationary GP.

\subsection{Datasets}
The obtained data were split into training and validation datasets. To test the performance of the model in advance, two additional waves with different wave heights and wave periods were generated. It should be pointed out that we did not randomly split the whole dataset into training and testing sets, which will lead to the over-performance of the predicting model for the time series task. Here, eight-fold cross-validation was applied to train the model, and then two different waves with new $H_s$ and $T_p$ will be used for elevating the real performance of the trained model.

For each case, a 30-minute discrete motion $\{x(t_i), i\in N\}$ and wave $\{\eta(t_i), i\in N\}$ were recorded. We processed the data as input--output pairs for learning as follows:
\begin{align}
	&\mathbf{X}^{n \times 2}_p = [x(t_{p-n}), x(t_{p-n}), \cdots, x(t_{p-1})]^T+[\eta(t_{p+w-n}), \eta(t_{p+w-n+1}), \cdots, \eta(t_{p+w+1})]^T \\
	&\mathbf{Y}^{m \times 1}_p = [x(t_{p}), x(t_{p+1}), \cdots, x(t_{p+m-1})]^T
\end{align}
where $t_p$ denotes the current time instant, $n$ is the time window expressed as the number of points of sequential wave and motion data used as inputs, $w$ is the wave lag expressed as the number of points of waves into the future, and $m$ is the prediction length expressed as the length of the prediction into the future. These definitions are consistent with those in Guo et al.~\cite{Guo2021}.

To reduce the variance of the parameters in our learning model, both the input waves and motions were also regularized as follows:
\begin{equation}
	\widetilde{x}(t_i) = \frac{x(t_i)-\text{A}}{\text{B}},
\end{equation}
where A and B are the mean and standard deviation of all available data in the training and validation datasets, which are constant in the present study. The 30-minute time history was first regularized, and then, it was scaled up to full scale. Finally, it was divided into input--output tensor pairs.

\begin{figure}[ht]
	\centering
	\includegraphics[width=0.75\linewidth]{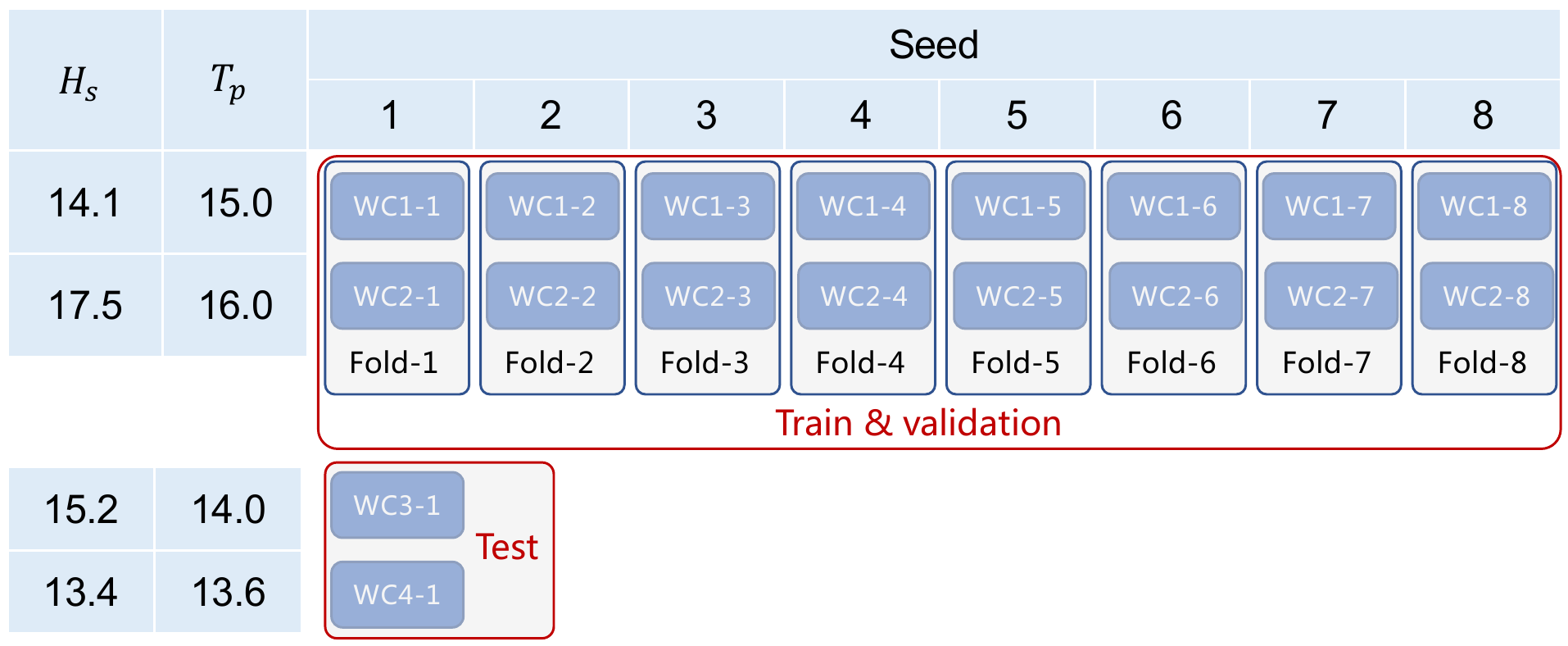}
	\caption{Illustration of the split of the training, validation, and testing datasets. Each case is a 30-minute wave test, and two cases become a fold. In total, there are eight folds for training and validation and one more fold with different wave parameters for testing.}
	\label{fig:dataset}
\end{figure}

As seen in Fig.~\ref{fig:dataset}, we have 18 cases in total with different incident wave time series. Sixteen of them were packed into eight folds for the k-fold training and validation process. Two other new cases were used for testing. The dataset doubled in size compared with those in our previous work (Guo et al.~\cite{Guo2021}), implying that the model with more learnable parameters could be trained for the learning task.

\section{Learning model}\label{sec:learningmodel}
\subsection{Deep learning algorithm}
The basic idea of a DL can be described as follows: for a pair of input tensor $\mathbf{x} \in \mathbb{R}^{r \times n}$ and output tensor $\mathbf{y} \in \mathbb{R}^m$, we have to find the parameters $\mathbf{p}$ (weights and bias) in a learning model $\mathscr{F}$ that performs nonlinear mapping $\mathscr{F}(\mathbf{x}; \mathbf{p})\approx \mathbf{y}$. Thereafter, the obtained prediction was compared with the ground truth through a loss function (mean square error, MSE) as follows:
\begin{equation}
	\mathscr{L}(\mathbf{y}, \mathbf{y}^*) = \sum_{i=1}^{N} \frac{\big|y_i - y^*_i\big|^2}{N},
\end{equation}

At the same time, the explained variance (EV) is defined as follows:
\begin{equation}
	\text{EV}(\mathbf{y}, \mathbf{y}^*) = 1 - \frac{\text{Var}[\mathbf{y}-\mathbf{y}^*]}{\text{Var}[\mathbf{y}]},
\end{equation}
where $\text{Var}[.]$ is the variance. The EV score is also used to describe the accuracy of the predictive time series. The best possible score is 1.0, and lower values are worse. 

By minimizing the MSE loss, the parameters in the learning model were updated. These steps were repeated until sufficient prediction accuracy was obtained. Once the parameters (weights and bias) were learned, the DL model $\mathscr{F}$ was ready for use.

An RNN is a generalization of the feed-forward NN with an internal memory. The basic idea is that the parameters in each cell are the same, and the output at the previous time step feeds the current cell as memory. The parameter size is independent of the length of the sequential input data, and with the memory flowing through the cells, the RNN provides a better performance for sequential input data. An LSTM model is a modified version of an RNN. The memory can easily spread through the cells by adding a direct shortcut between cells. Furthermore, LSTMs are well-suited to classify, process, and predict time series given time lags of unknown durations. The detailed formulations of an LSTM cell can be found in~\cite{Hochreiter1997}.

\begin{figure}[ht]
	\centering
	\includegraphics[width=\linewidth]{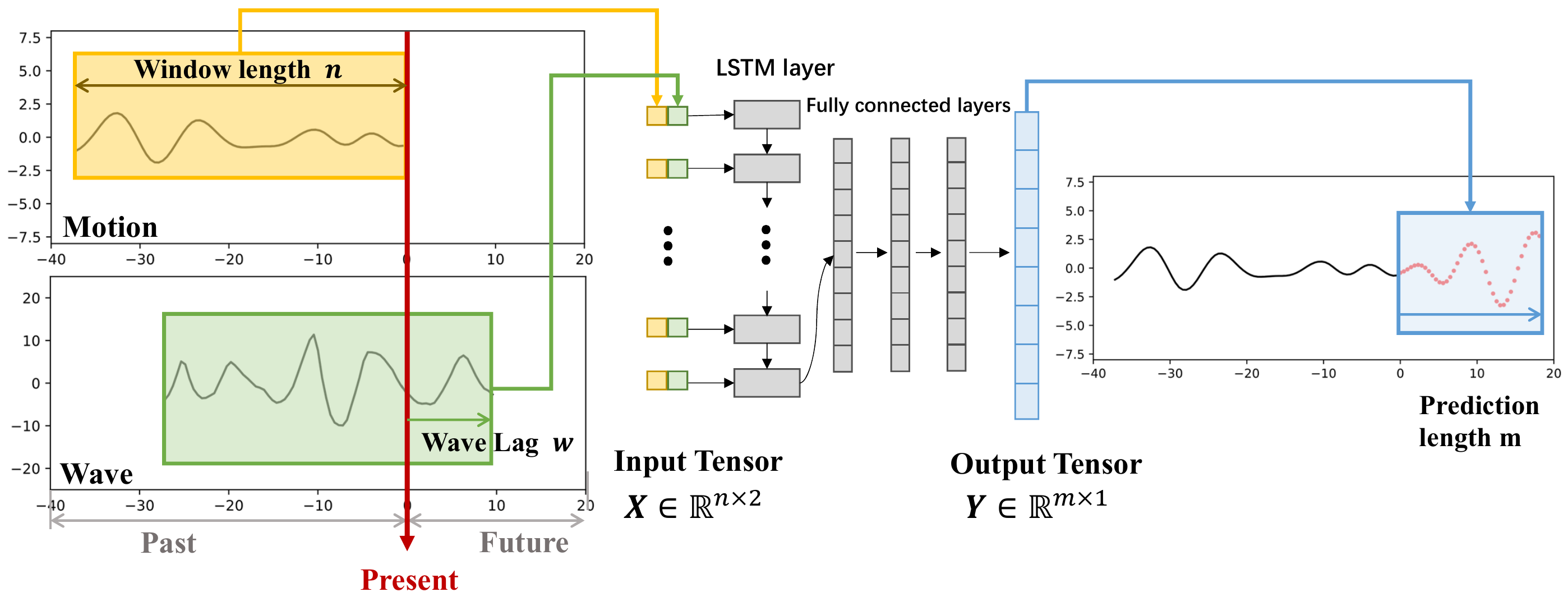}
	\caption{Illustration of the LSTM based DL model structure for an offshore platform motion prediction proposed in~\cite{Guo2021}.}
	\label{fig:lstm-model}
\end{figure}

Guo et al.~\cite{Guo2021} proposed an LSTM based DL model for predicting the motions of an offshore platform in waves. As shown in Fig.~\ref{fig:lstm-model}, the input data are first fed into LSTM cells. Then, the output tensor of the last LSTM cell goes through several FC layers with a hyperbolic tangent ($\tanh$) as the activation function. In the present work, we followed this specification to build the present model.

\subsection{Dropout for uncertainty estimation}
\begin{figure}[ht]
	\centering
	\includegraphics[width=0.65\linewidth]{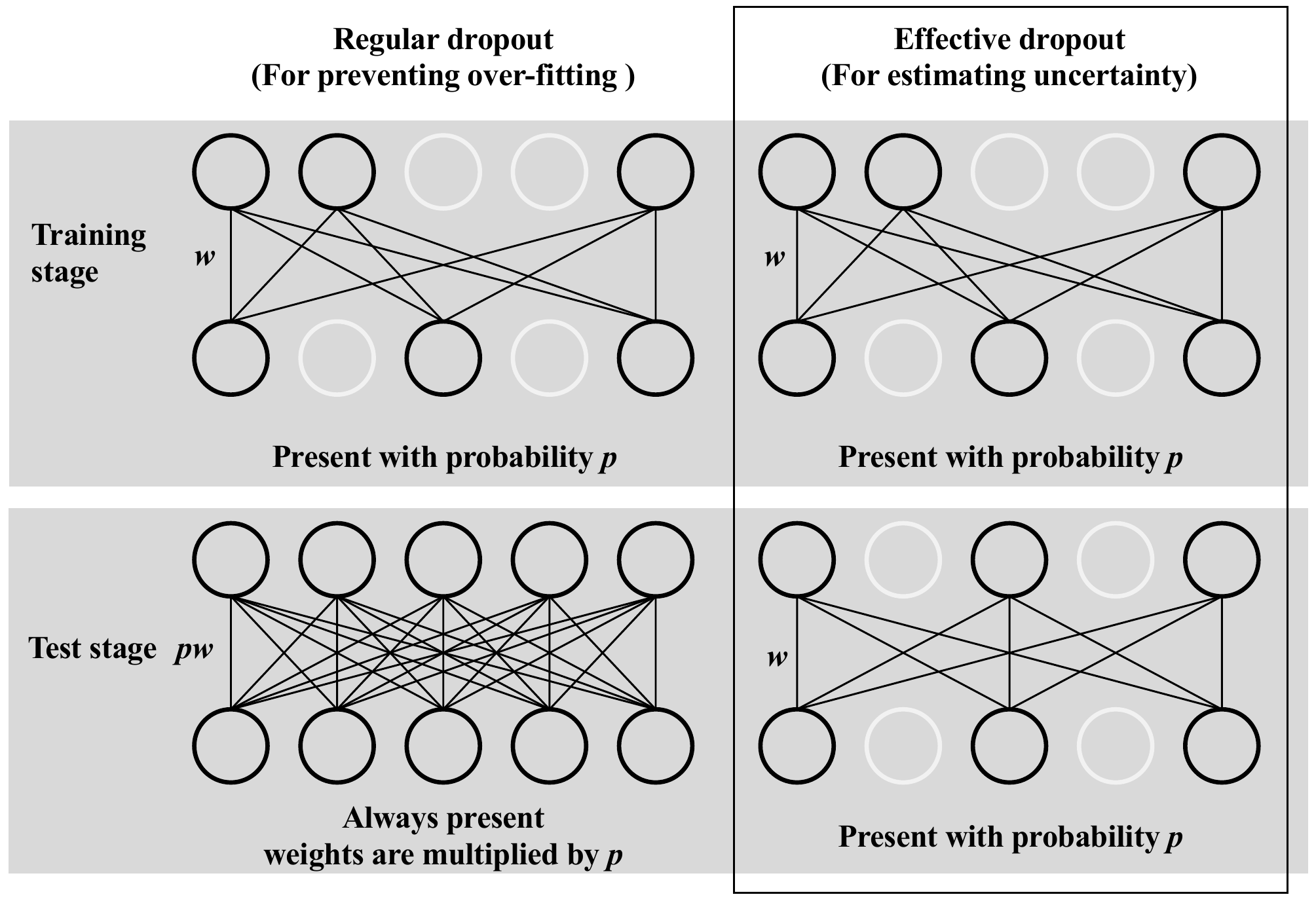}
	\caption{Illustration of regular dropout procedure for preventing the over-fitting problem (left-hand column), and the effective dropout for estimating the uncertainty in the right-hand column.}
	\label{fig:dropout}
\end{figure}

Dropout, which refers to dropping out units at hidden layers, is a well-used technique to prevent the over-fitting problem in an NN\cite{Srivastava2014}. As shown in Fig.~\ref{fig:dropout}, at the training stage, a random fraction, $p$, of nodes is ignored for each iteration for each training sample. For a regular use of dropout, when the training process concluded, all the nodes are effective but reduce the weight by a factor of p to make outputs in the test phase. Dropout helps the model to learn more robust features that are useful in conjunction with many random subsets of the other neurons.

\begin{figure}[ht]
	\centering
	\includegraphics[width=\linewidth]{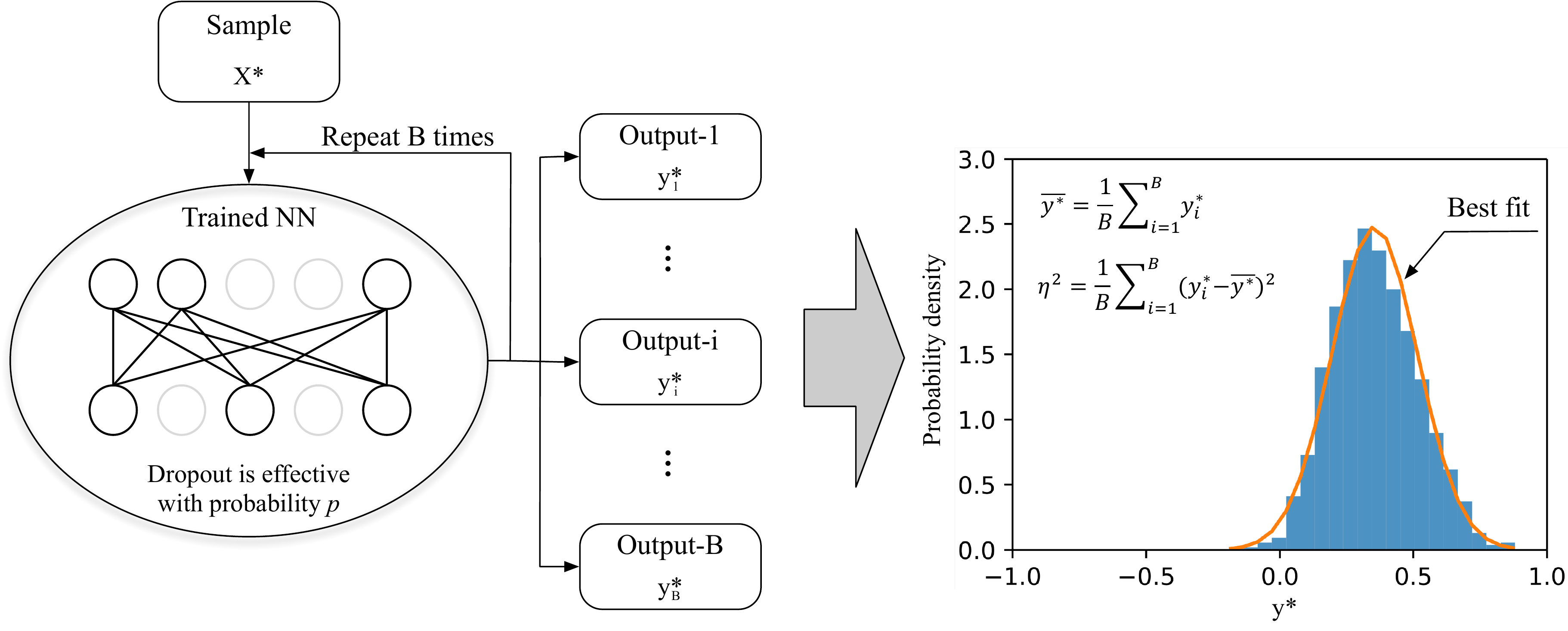}
	\caption{Dropout for uncertainty estimation. For one single input, the inference repeats B times, and the mean and standard deviation are calculated based on the outputs.}
	\label{fig:dropoutBtimes}
\end{figure}

With effective dropout in the testing stage, for each inference, the result varies for a single input tensor. We assumed that these outputs followed a Gaussian distribution $y^* \sim \mathcal{N}(\mu,\,\sigma^{2})$, and then, the mean $\mu$ could be regarded as the prediction and variance $\sigma^{2}$ as the uncertainty information. As seen in Fig.~\ref{fig:dropoutBtimes}, the dropout was still effective during the testing stage. We repeated the inference $B$ times to have $B$ different outputs. Then, instead of a deterministic result, a distribution of the predictive time series is obtained.

Training a DL model with effective dropout, in a sense, is equal to training a collection of thinned NNs with extensive weight sharing (with $n$ units, we have $2^n$ thinned models). Here, as suggested by Gal and Ghahramani~\cite{Gal2016}, the $2^n$ possible thinned models could be treated as Monte Carlo samples. At each time, the output is made by a randomly selected model from $2^n$ possible models, leading to a varying prediction for one single input sample. Then, the confidence of this single prediction was evaluated based on the obtained result collection $[y^*_1, y^*_2, ..., y^*_B]$. This offers a handy way to estimate uncertainty with minimal changes in most existing DL models. 

\begin{algorithm}[H]
	\SetAlgoLined
	\KwIn{Input data $x$, DL model $\mathscr{F}$ with dropout probability $p$, number of iterations $B$}
	\KwOut{Prediction mean $\overline{y^*}$ and its variance $\sigma^{2}$}
	\For{i = 1 to B}{
	  $y^*_i = \mathscr{F}(x, p)$
	 }
	$\overline{y^*} = \frac{1}{B}\sum_{i=1}^B y_i$\;
	$\sigma^{2} = \frac{1}{B}\sum_{i=1}^B (y_i - \overline{y^*})^2$
	\caption{Effective dropout for uncertainty estimation}
\end{algorithm}

\subsection{DL model}
\begin{figure}[ht]
	\centering
	\includegraphics[width=0.8\linewidth]{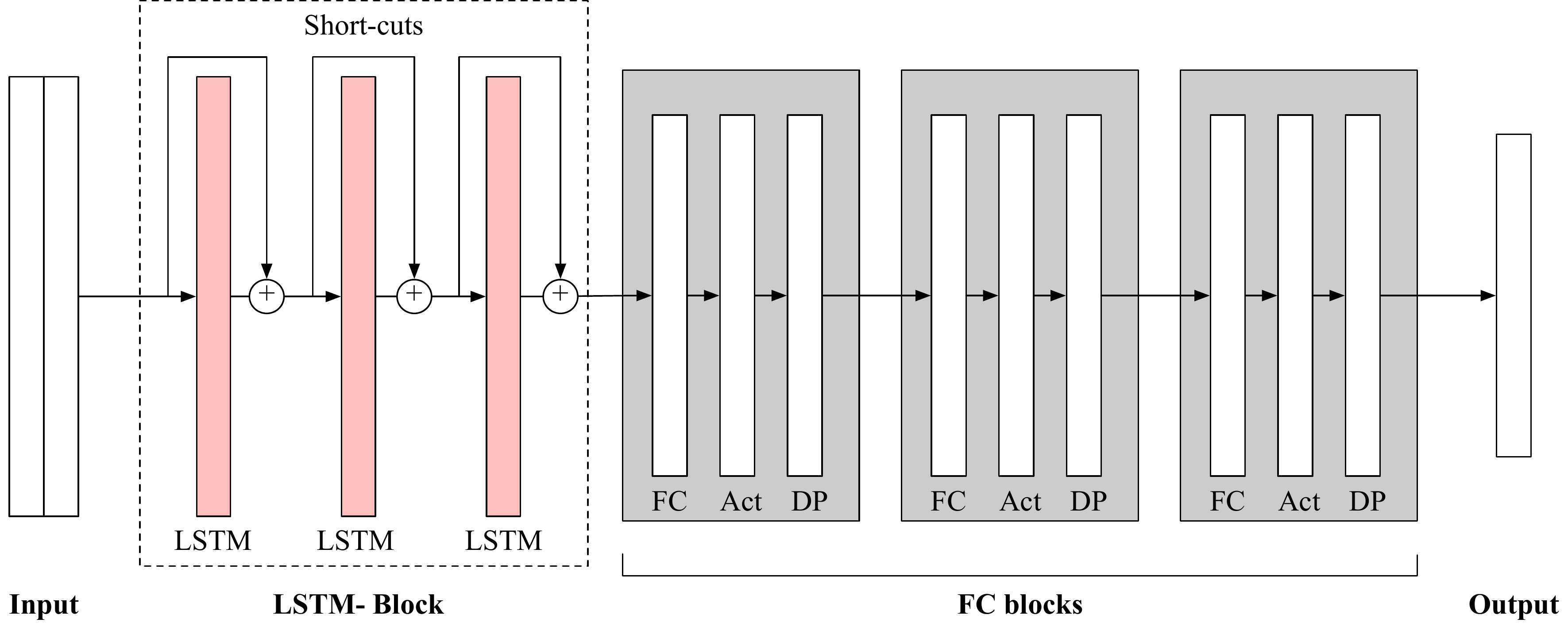}
	\caption{Illustration of the DL model with dropout layers for the uncertainty estimation of the prediction. FC: fully connected, Act: activation, DP: dropout.}
	\label{fig:model-lstmdropout}
\end{figure}

In this study, for uncertainty estimation, we inserted the dropout layers into the original model. The model structure is illustrated in Fig.~\ref{fig:model-lstmdropout}. Dropout performs regularization on the model and also helps to prevent the over-fitting problem. As we doubled the dataset compared to the study presented in \cite{Guo2021}, the model used here had more trainable parameters. In LSTM blocks, shortcut connections were also added between LSTM layers, and this allows the model to skip some LSTM layers to avoid the vanishing gradient problem.

The model parameters are listed in Tab.~\ref{tab:basicmodel}. The model predicts motions $m$ steps into future, and the wave lag and time window is set to be $m$ and $3m$, respectively, as suggested in~\cite{Guo2021}. In this study, the prediction length $m$ varied from 20 to 80. For this learning task, the model had three LSTM layers, and five FC blocks with fifty neurons in each FC layer.

\begin{table}[htp]
	\caption{Hyper parameters used for the present model.}\label{tab:basicmodel}
	\begin{tabular}{lc|lcl}
	\toprule
	Parameter             & Value & Parameter                     & Value     \\ 
	\midrule
	Forward step         & m     & Learning rate                 & 0.01       \\
	Wave lag             & m     & Learning rate schedule        & step decay \\
	Time window          & 3m    & Number of LSTM layers         & 2          \\
	Motion               & Heave & Number of LSTM output neurons & 200        \\
	Max number of epochs & 200   & Number of FC blocks           & 5          \\
	Early stopping       & True  & Number of neurons in FC layer & 80         \\
	Batch size           & 2048  & Activation function           & tanh       \\
	Optimizer            & Adam  & Dropout probability           & 0.315      \\ 
	\bottomrule
	\end{tabular}
\end{table}

For training the model, early stopping criteria were applied to prevent over-fitting, and the Adam algorithm~\cite{Kingma2014} was used to minimize the loss. Mini-batch gradient descent was also applied with a batch size of 2048. As seen in Fig.\ref{fig:dataset}, all available data cases were split into nine folds. Each fold contains two cases with different combinations of wave heights and wave periods. An eight-fold cross validation was applied to train the model. The last fold was used to test the performance only. A step-decay algorithm is used as the learning rate schedule. The initial learning rate was set at 0.01 for the first 10 epochs, and then it was decayed with a rate of 0.1 at every 50-epoch milestone. The model was trained on a PC with an NVIDIA GeForce RTX-3090 graphics card. A code demonstration can be found online (\url{https://github.com/XiaoxG/waveMotion-lightning}).

\section{Results and discussion}\label{sec:results}
\subsection{Predictions with effective dropout}
\begin{figure}[ht]
	\centering
	\includegraphics[width=\linewidth]{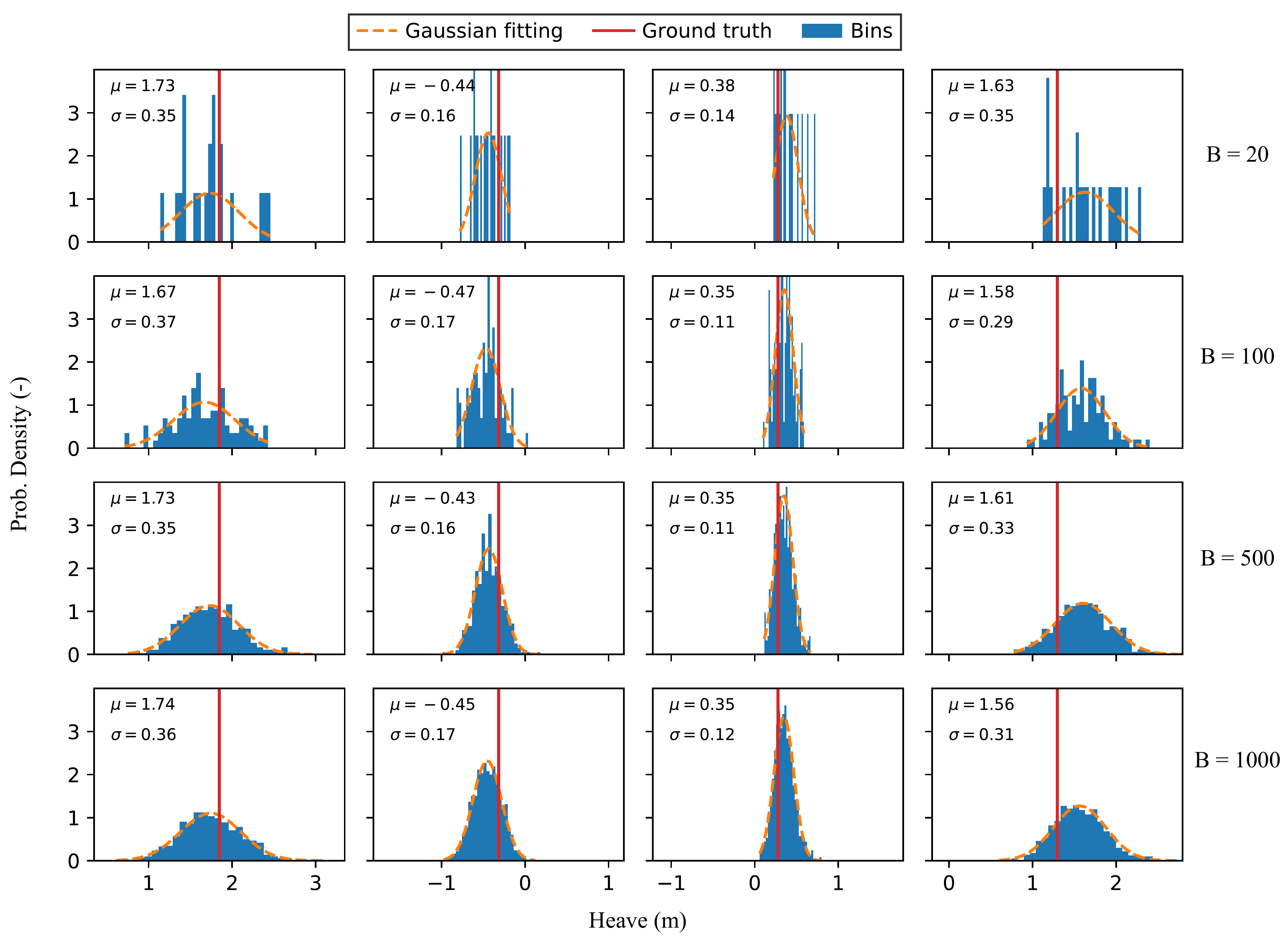}
	\caption{Probability densities of the $B$ times outputs for a single point. In each column, the predictions are made at the same point. The dashed line indicates the best Gaussian distribution fitting based on the outputs, and the red solid line indicates the ground truth. $\mu$: mean, $\sigma$: standard deviation.}
	\label{fig:histBs}
\end{figure}

As the dropout is still effective during the testing stage, we can obtain B different outputs for one single input if we repeat the prediction B times. In Fig.~\ref{fig:histBs}, we randomly selected four predictions from the test dataset. The probability density of these point predictions is shown with increasing repetition times B as columns. It is seen that the probability density of the single-point-prediction clearly follows a Gaussian distribution. Then, we use mean $\overline{y^*}$ and standard deviation $\sigma$ to describe the distribution of the B times prediction. 

As seen in Fig.~\ref{fig:histBs}, $B=500$ is adequate to get a very good probability density function in a Gaussian distribution form. The 90\% confidence interval (CI) of the prediction was also estimated as follows:
\begin{equation}
	\text{CI}_{90\%} = y^* \pm 1.645 \sigma.
\end{equation}

\begin{figure}[ht]
	\centering
	\includegraphics[width=0.9\linewidth]{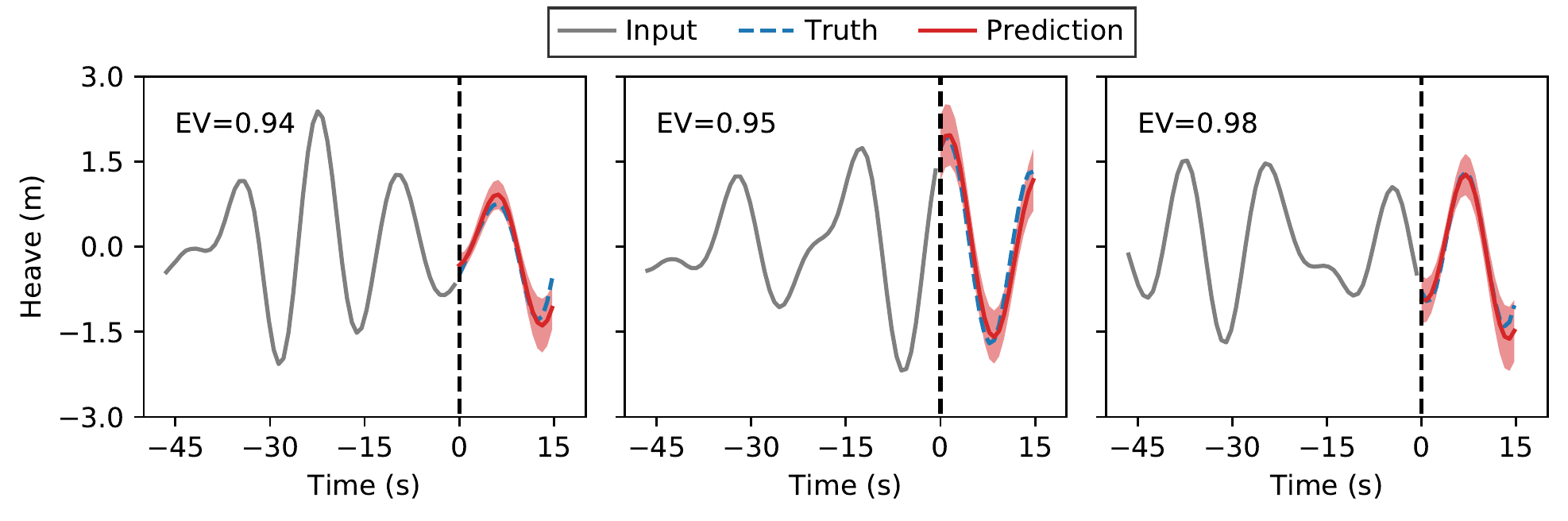}
	\caption{Examples of time series of the prediction with 90\% confidence level made on the test dataset. The three time series examples were randomly selected from all the results in the test dataset. The red solid line represents the predictive mean, and the shaded area represents the 90\% confidence region. EV: explained variance.}
	\label{fig:Timehis_CI}
\end{figure}

Figure~\ref{fig:Timehis_CI} shows six time series of the prediction with a prediction length of 20 points (approximately 14.4 seconds) into the future. The red shaded area represents the 90\% CI of the prediction. The present results were randomly selected from the test dataset, which showed well-agreed results. It should be pointed out that the test datasets were obtained based on two new wave conditions. The learned model predicted the occurrence of the next crest or trough (changing point) with very good accuracy. However, the amplitude varies in a relatively larger range. The 90\% CI of predictions at crests and troughs are wider than those of other points. The present model showed a strong ability to predict the occurrence of the next crest or trough.

It should be noticed that the prediction still has a relatively large uncertainty at the first few points of the predicted time series. From a general point of view, the prediction at points very close to the known points should have a high accuracy. However, the fact is that the prediction confidence was not improved at the first few points. Another feature is that the predicted time series is very smooth. Therefore, we concluded that instead of predicting the time series from the close to the distant, the model learned the period at first, and then fit the curve by minimizing the overall differences on the prediction time series in a point-wise context. It seems that the model learned a kernel at first, and then used the kernel to find a smooth curve that had the lowest loss; this is similar to GPR.

\begin{figure}[ht]
	\centering
	\includegraphics[width=0.9\linewidth]{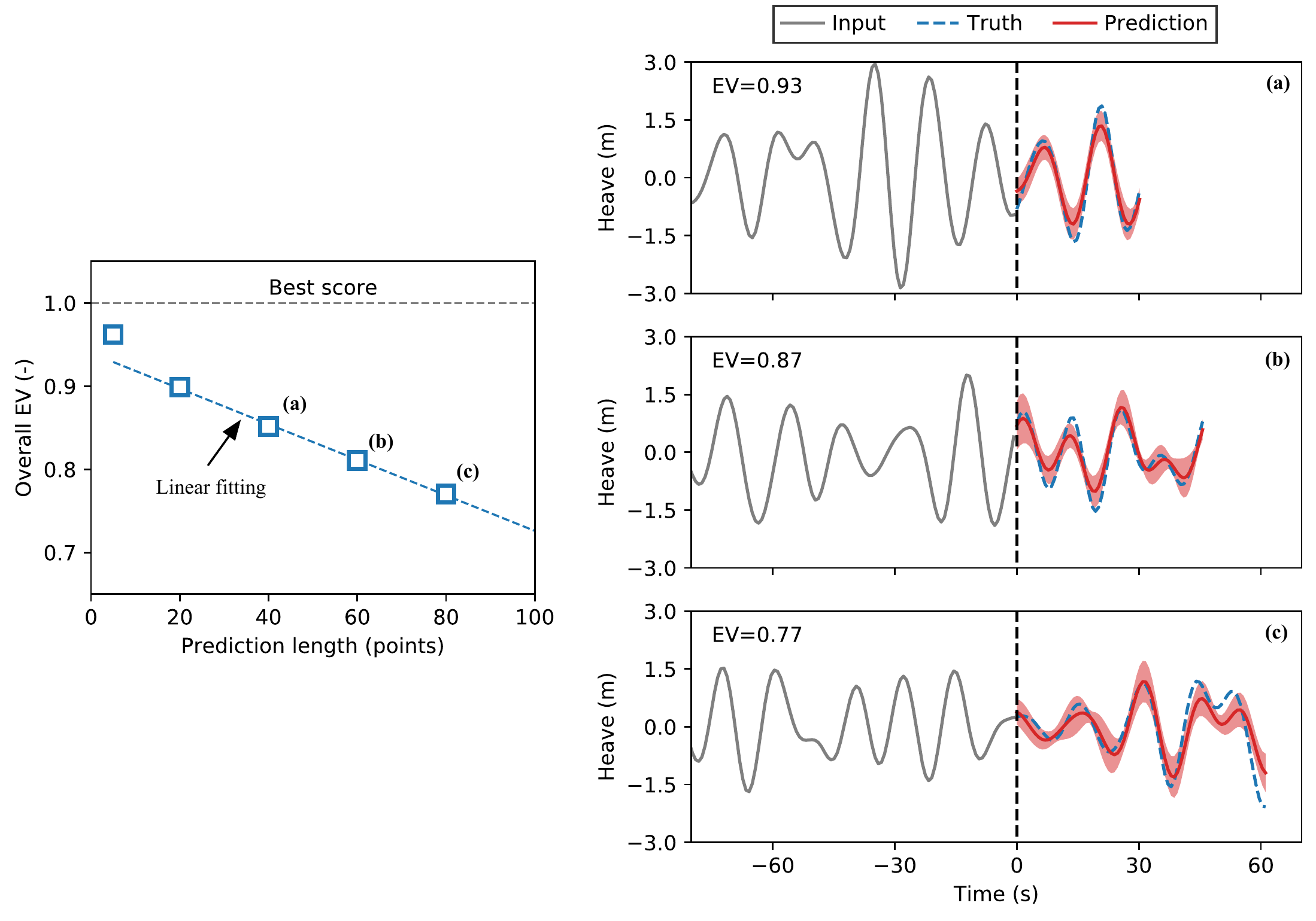}
	\caption{Left: The overall averaged EV score on test dataset with increasing prediction length; Right: Examples from the prediction time series with increasing forward step from 40 to 80 points. The red solid line represents the predictive mean, and the shaded area represents the 90\% confidence region.}
	\label{fig:series_manylenth}
\end{figure}

By keeping the ratio between the wave lag and forward step $w/m=3$, the prediction could extend more points into the future, which was consistent with Guo et al.~\cite{Guo2021}. As seen in Fig.~\ref{fig:series_manylenth}, the performance of the model was still good when the forward steps extended to 80 points (EV > 0.7). For all the prediction models, the hidden layers (LSTM and FC layers) of the model were consistent as listed in Tab.~\ref{tab:basicmodel}. With the same model (the same number of learnable parameters), the overall averaged EV score on the test dataset with prediction length from 20 to 80 points linearly decreased. As seen in fig.~\ref{fig:series_manylenth}, with increasing forward steps, the periodicity of the outputs is still consistent with the ground truth, but the predictive amplitude got worse.
 
\subsection{Covariance of the predictions}
\begin{figure}[ht]
	\centering
	\includegraphics[width=0.55\linewidth]{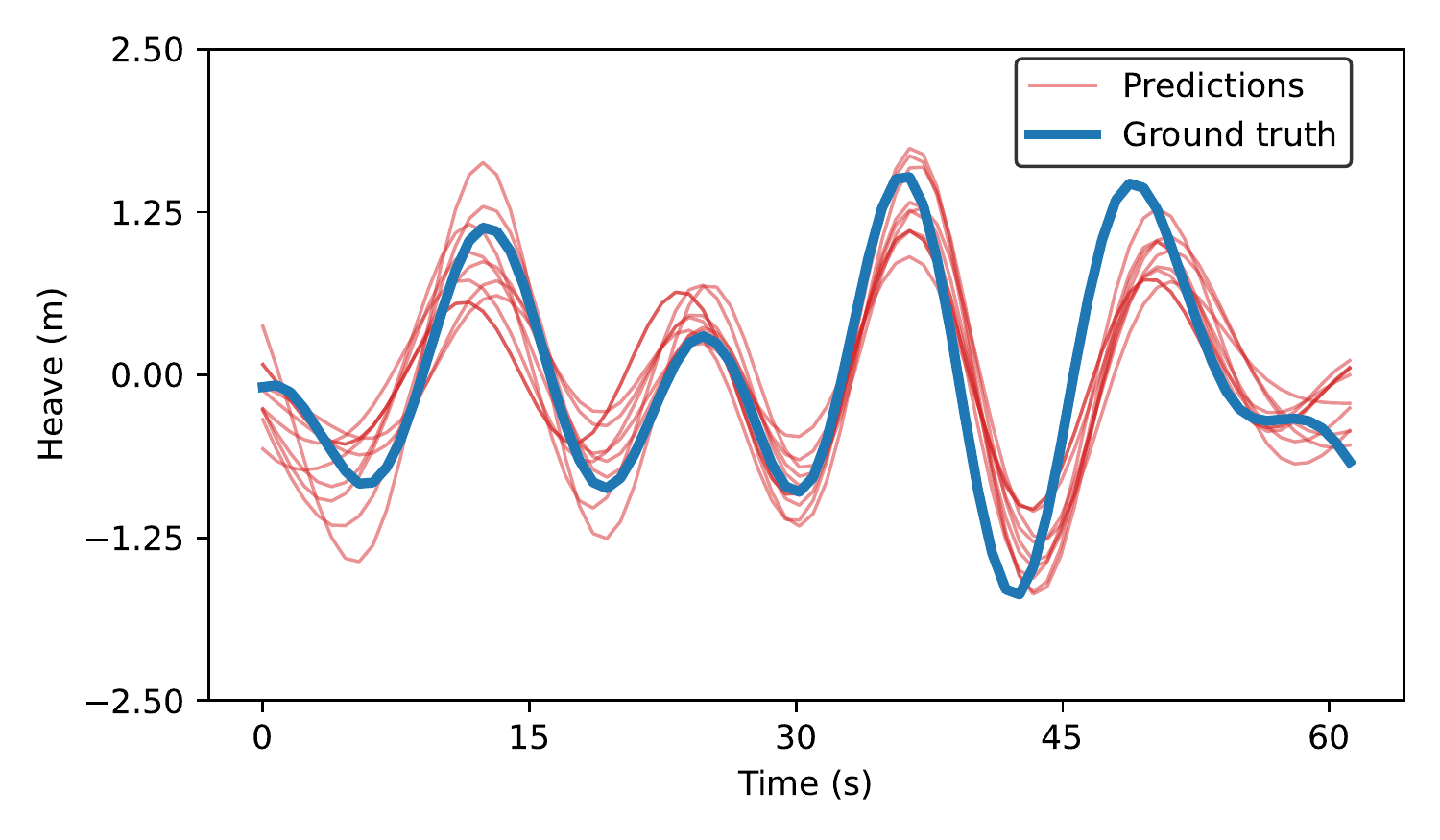}
	\caption{Time series of 10 possible predictions made by the learned model from the same input with the corresponding ground truth (blue line). The prediction length is 80 points into the future.}
	\label{fig:pos_preds}
\end{figure}

Figure~\ref{fig:pos_preds} show 10 possible predictions made by the learned model by one single input and the corresponding ground truth. The prediction length is 80 points ahead. The 10 predictions varied in amplitude in a point-wise context, but they had approximately the same varying periods. As discussed above, all the possible predictions were obtained via a DL model with effective dropout. For any single point prediction, the outputs from the DL model followed a Gaussian distribution (see Fig.~\ref{fig:histBs}). Consequently, this collection of the predictions made for one input could be regarded as a GP.

As defined by Williams and Rasmussen~\cite{Williams2006}, a GP is a collection of random variables, any finite number of which has a joint Gaussian distribution. A GP $\mathcal{G}\big(m(x), k(x, x')\big)$, which is completely defined by its mean and covariance function, could be used to describe the collection of motion time series predicted by the present DL model. Here, we used the obtained 10 80-point-predictions, as shown in Fig.~\ref{fig:pos_preds}, as an example to investigate its properties from a view of GP. The covariance matrix was calculated as follows:
\begin{equation}
	\text{COV}_{i,j}(X_i, X_j) = \mathbb{E}\big[(X_i- \mathbb{E}[X_i])(X_j- \mathbb{E}[X_j])\big].
\end{equation}

\begin{figure}[ht]
	\centering
	\includegraphics[width=\linewidth]{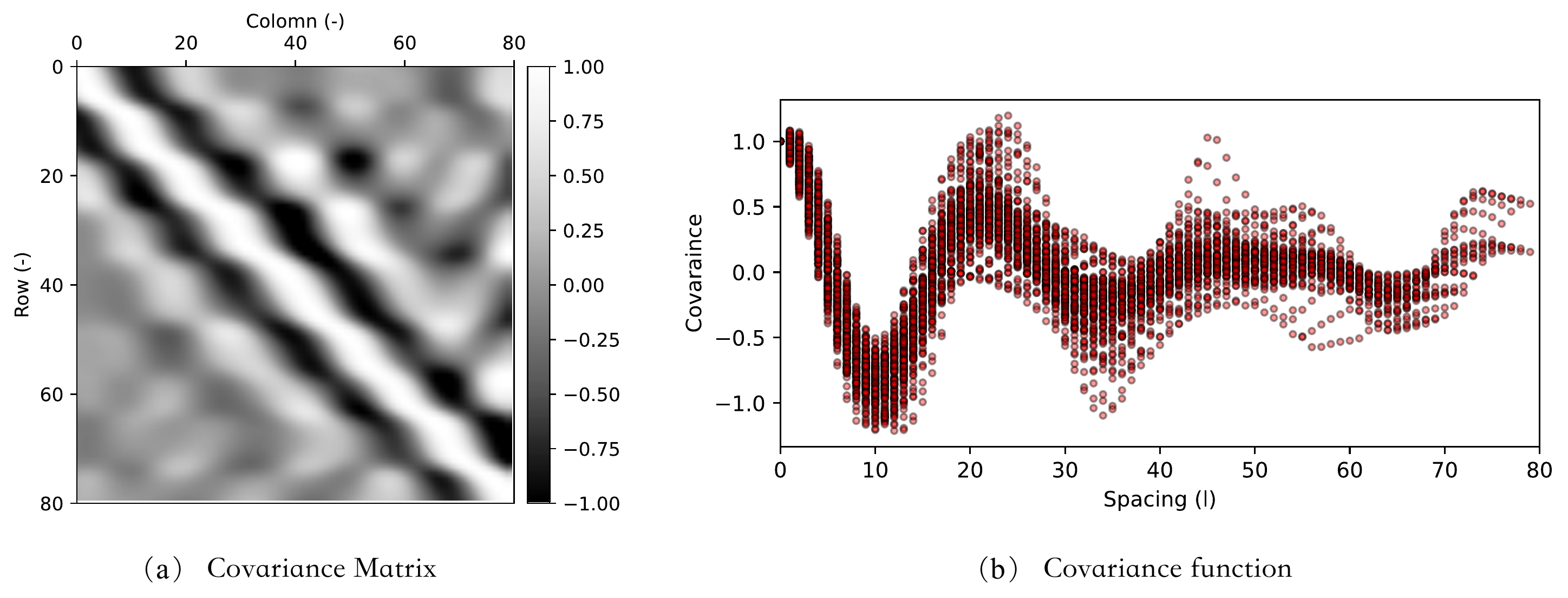}
	\caption{(a) Visualization of the covariance matrix and (b) covariance as a function of the spacing $l$. The matrix came from the collection of 80-point-predictions for one single input, as shown in Fig.~\ref{fig:pos_preds}.}
	\label{fig:covshow}
\end{figure}

The covariance matrix for the collection of predictions, as shown in Fig.~\ref{fig:pos_preds}, is depicted in Fig.~\ref{fig:covshow} (a). To have a better view, each row in the covariance matrix was normalized by the corresponding diagonal term. It is seen that the terms close to the diagonal have large values (close to 1), and the covariance matrix is symmetric. Evidently, in a back diagonal direction, it is periodic. The covariance as a function of the spacing $l$ is shown in Fig.~\ref{fig:covshow} (b). The period can be identified as approximately 20 points in length. It is seen that when the spacing is less than 10 points, the covariance only depends on the spacing $l = |x-x'|$. 

\subsection{Prediction with input uncertainty}
\begin{figure}[ht]
	\centering
	\includegraphics[width=\linewidth]{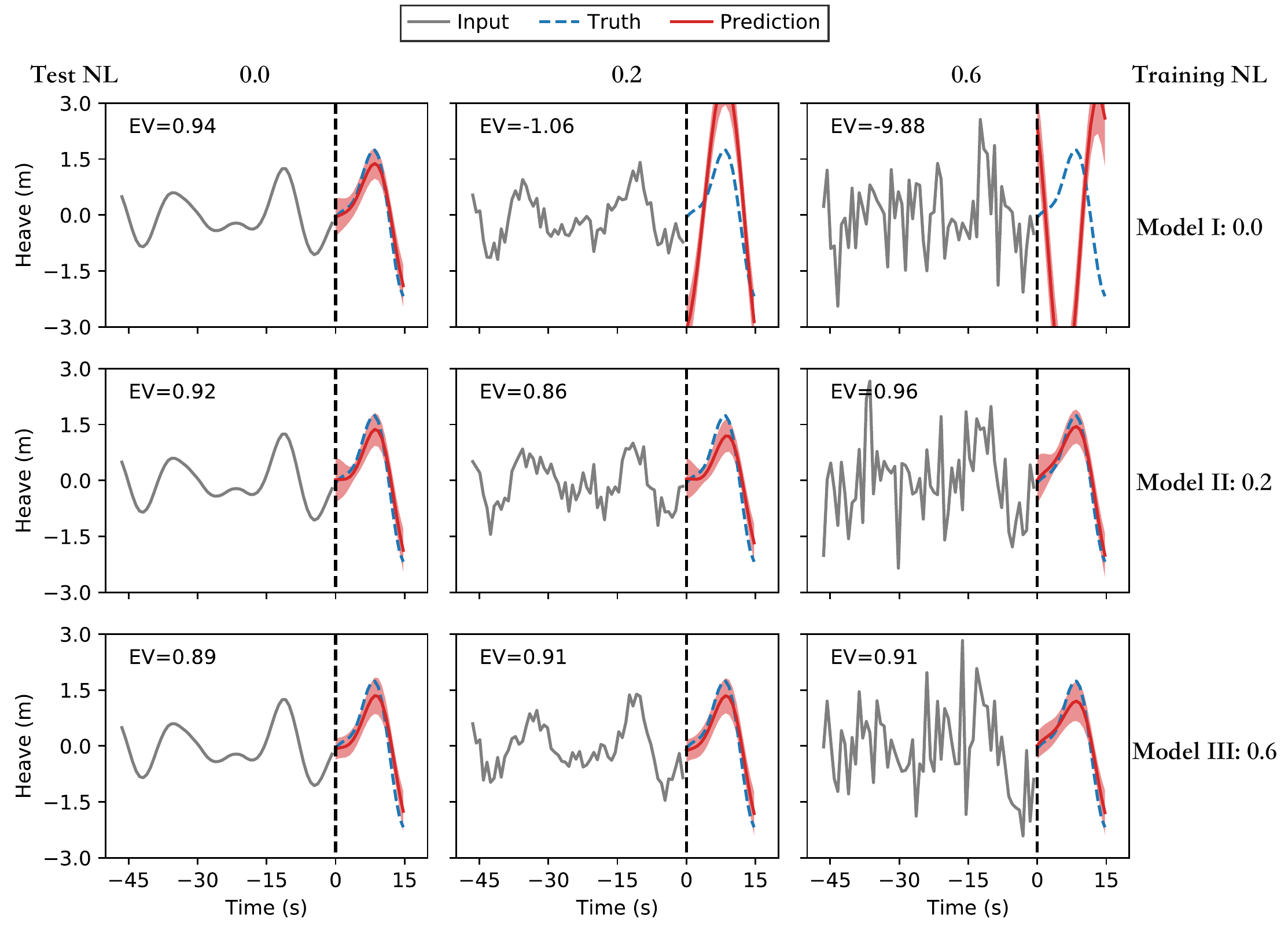}
	\caption{Time series of the 20-point-forward prediction with a 90\% confidence level made on the same input. In each row, the learned model was fed with data containing different levels of noise. In each column, with the same level of noise in test data, the learned model was trained with different noisy data. The red solid line represents the predictive mean, and the red shaded area represents the 90\% confidence region. The EV indicates the explained variance score, while NL represents the noise level.}
	\label{fig:timeseries_nl}
\end{figure}

In reality, the observations (input) are always noisy. In this section, we assumed that the observations take the form $x = x + \epsilon$, where x is the noise-free data and $\epsilon \sim \mathcal{N}(0,\,\sigma_n^{2})$ is an independent and identically distributed noise contribution. Here, $\sigma_n$ is defined by the percentage of the overall standard deviation of the heave motion in the test dataset. We directly used the learned model without any knowledge of noise at first, similar to the trained model used in the previous section. The model failed with noisy input as seen in the first row in Fig.~\ref{fig:timeseries_nl}. 

Then, by adding noise into the training data, the model knows which information is important for the prediction task. It should be pointed out that the noise was only added to the input data, while the ground truth was kept clean. Two models were trained with a noise level of 0.2 and 0.6, and then tested on the same input with different level of noises, respectively. With the knowledge of noise, the model gives good predictions on test inputs, which means that the model learns how to handle the noise, even if noise level is much higher than that in the training data.

\begin{figure}[ht]
	\centering
	\includegraphics[width=0.65\linewidth]{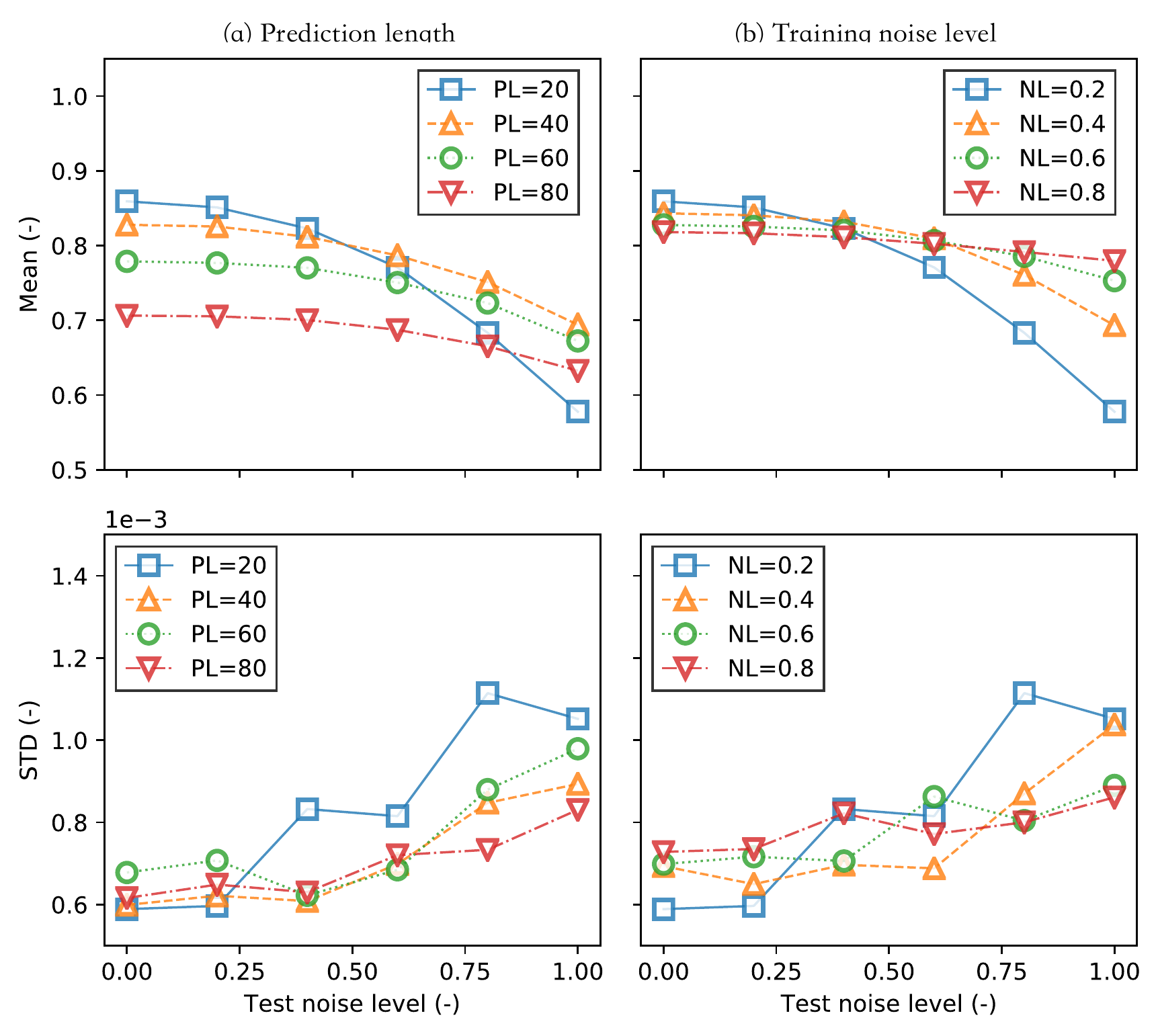}
	\caption{The mean and standard deviation of the EV score on the test dataset with varying noise levels from 0.0 to 1.0 for different prediction lengths with a training noise level of 0.2 in the left-hand column and different noise levels in the training dataset with a prediction length of 20 in the right-hand column. The noise level is the ratio between the standard deviation of the noise and the heave motion. PL indicates prediction length, and NL represents noise level.}
	\label{fig:evscore_noisy}
\end{figure}

In Fig.~\ref{fig:evscore_noisy}, we evaluated the models in terms of the overall averaged EV score on the whole test dataset. For all prediction lengths, the performance of the prediction dropped with increasing test noise level. In general, a shorter prediction length and lower test noise level mean better performance. However, for the cases with a noise level higher than 0.75, the performance of the prediction with 20 points forward got worse more quickly than others. Nonetheless, if we increased the noise level in training data, the performance on the test data with higher level of noise became better. Meanwhile, the performance on the test data with less noise was still good. Based on this perspective, we should add noises into training data to improve the overall performance of the model. 

\section{Conclusions}
In this study, with the help of dropout layers, we extended the DL model proposed in Guo et al.~\cite{Guo2021} to predict future heave motions of an offshore platform with uncertainty estimation. With the proposed structure, the DL model predicted the future heave motions with a very good accuracy compared to ground truth. By repeating the inference several times, the collection of the output time series was a GP. Since inserting dropout layers into the existing DL model is very easy, we provided a handy way to estimate the prediction uncertainty of existing DL model. We believed that the DL model with dropout learned a kernel inside, and the procedure was equivalent to a GPR. Finally, adding noise into training data could help the model to learn generalized features from the input information, thereby leading to a better performance on test data with a wide noise level range.

The proposed DL model in this study showed a very strong ability to predict motions of an offshore platform. The amounts of data are a prerequisite for this type of learning process. The next question entails whether the model could be pre-trained on a large training dataset and then used for a particular platform with a very limited dataset. Another issue is that only waves are considered as environmental excitations; thus, what happens if wind and current effects are taken into account? These will be the future work of the authors.

\section{Acknowledgements}
This study was supported by Major Science and Technology project of Hainan Province (ZDKJ2019001), Shanghai Sailing Program (Grant No. 20YF1419700) and State Key Laboratory of Ocean Engineering (Shanghai Jiao Tong University) (Grant No. 1915).

\bibliographystyle{unsrt}


\end{document}